\documentclass{article} 

\usepackage{iclr2027_conference, time}

\iclrfinalcopy

\usepackage{amsmath,amsfonts,bm}

\def\eqref#1{equation~\ref{#1}}

\def\1{\bm{1}}

\DeclareMathAlphabet{\mathsfit}{\encodingdefault}{\sfdefault}{m}{sl}
\SetMathAlphabet{\mathsfit}{bold}{\encodingdefault}{\sfdefault}{bx}{n}

\usepackage{hyperref}
\usepackage{url}
\usepackage{graphicx} 
\usepackage{xcolor}
\usepackage{booktabs}
\usepackage{amsmath}
\usepackage{multirow}
\usepackage{algorithm}
\usepackage{algorithmicx,algcompatible}
\usepackage{algpseudocode}
\usepackage{array}
\usepackage{subcaption}
\title{Compress to Remember: Learning Compact Memory via On-Policy Distillation for Long Video Generation}

\author{%
Xiaoyu Wu$^{1, *}$ \quad
Weihang Guo$^{2, *}$ \quad
Yifei Wang$^{2, *}$ \quad
Xinze Feng$^{1}$ \\[0.55em]
\textbf{Lydia E. Kavraki$^{2}$ \quad
Zhiwei Steven Wu$^{1}$} \\[0.4em]
$^{1}$Carnegie Mellon University
\qquad
$^{2}$Rice University \\
\texttt{xiaoyuwu@andrew.cmu.edu, wg25@rice.edu, yw251@rice.edu,  } \\
\texttt{xinzef@andrew.cmu.edu, kavraki@rice.edu, zstevenwu@cmu.edu} \\
\small $^{*}$Equal contribution.
}

\usepackage[normalem]{ulem}
\definecolor{addblue}{HTML}{4A86C8}

\algnewcommand\Input{\item[\textbf{Input:}]}
\algnewcommand\Output{\item[\textbf{Output:}]}
\newif\ifshoworiginal
\showoriginaltrue

\begin{document}
\maketitle
\lhead{Under Review}

\begin{abstract}
Standard video generators do not natively compact historical context into
reusable memory tokens. As generation continues, the growing history makes
it increasingly difficult to retain information from earlier frames due to long-context degradation. Key-frame-based approaches address this challenge
by retaining selected past frames, but can discard information needed for
future generation. Rather than relying on frame selection alone, we study
whether a frozen video generator can supply the supervision needed to
learn a compact representation of the history. We propose Prediction-Aligned Context Compaction (PACC), which uses a learned
compressor to aggregate information across past frames into compact
memory tokens. 
We train the compressor through on-policy distillation, using the same
frozen generator both as a student when conditioned on compressed memory
and as a teacher when conditioned on the full history. The student
generates continuations, while the teacher provides targets for the
same noisy inputs at each denoising step. Only the compressor is
updated to align the student's predictions with these targets.
We evaluate PACC on MBench, which jointly measures memory-event coverage
and consistency. PACC outperforms the strongest baseline by 6.63 points on Causal-rCM and 3.19 points on Causal Forcing. Evaluation on VBench-Long using MovieGen prompts further shows that PACC produces minute-long videos with generation quality competitive with baselines.
Together, these results show that learning to compact historical context
can improve long-video memory without modifying the underlying generator.
\end{abstract}

\section{Introduction}
\label{sec:introduction}

Generating long videos is computationally demanding because each frame
contains many spatial tokens, and the temporal context grows as generation
proceeds.  Even after latent compression, approximately five seconds of video at
$832\times480$ resolution in our setting occupy 33,000 tokens---a sequence
budget corresponding to roughly 23,000 words of English
text.\footnote{Our setup uses 21 latent frames with 1,560 spatial tokens
each. The text estimate uses the WebText2 ratio of 1.4 tokens per word
reported by \citet{kaplan2020scaling}.} These costs make training on long sequences expensive, motivating
autoregressive approaches that extend models trained on short clips to
longer durations~\citep{causal-forcing, self-forcing, causal-rcm}. However, generating successive chunks requires deciding
how to preserve useful information from the growing history. Retaining all
past tokens becomes increasingly costly and can lead to degradation as
the context extends beyond the model's training horizon, while maintaining
a sliding window can remove information needed to maintain consistency.

\begin{figure}[t]
\vspace{-0.15in}
    \centering
    \includegraphics[width=\linewidth]{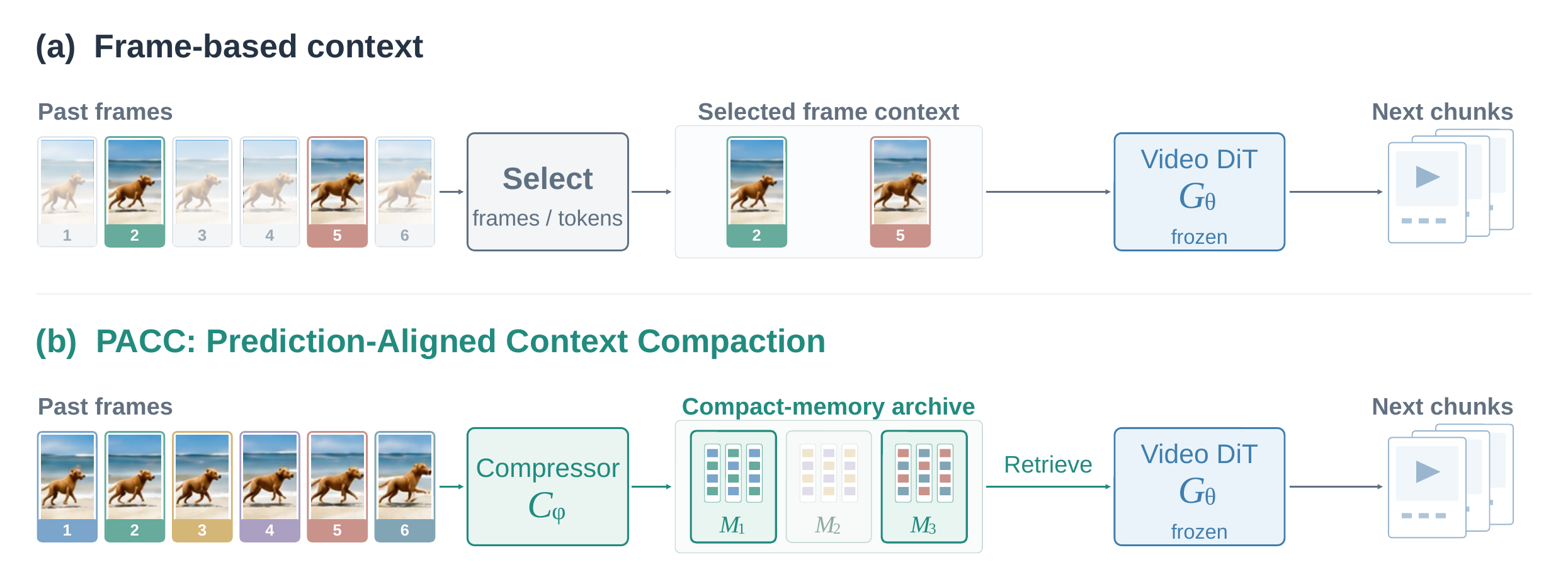}
    \caption{\textbf{Frame-based context versus learned memory compaction.}
    (a) Frame-based approaches select existing historical frames or tokens
    to condition future generation.
    (b) Prediction-Aligned Context Compaction (PACC) compacts each completed
    video block into learned memory tokens, stores them in an archive, and
    selectively retrieves compressed blocks within a bounded active context.
    Highlighted memory groups are retrieved; faded groups remain archived.
    Compaction changes the representation available for retrieval rather than
    eliminating selection.}
    \label{fig:pacc-memory-compaction}
\end{figure}

Language-model agents address related context constraints through context
compaction: summarizing accumulated interactions and continuing from the
resulting summary~\citep{context-compaction}. Related work also explicitly
trains language models to produce reusable compressed prompt
representations~\citep{gist-tokens}. These operations rewrite historical
information into new compact tokens. By contrast, many long-video
generation approaches manage history by retaining a subset of existing
tokens through sliding windows or key-frame
selection~\citep{yang_2025_longlive,relax-forcing,
chen_2026_pyramid,hu_2026_longliverag,ye_2026_dysink}.
This distinction raises a direct question:
\emph{Can a frozen video generator teach a compressor to produce compact
memory that preserves its predictions under full-history conditioning?}

Video makes this question particularly compelling because of its
substantial spatial and temporal redundancy. Within a frame, large regions
often share similar colors or textures. Across consecutive frames, the
same subjects and backgrounds persist with only small changes. This
redundancy creates opportunities to compact history beyond selecting
individual observations. Rule-based or similarity-based frame selection
and sliding windows discard unselected content, potentially losing
information needed for future generation. As illustrated in
Figure~\ref{fig:pacc-memory-compaction}, a learned compressor can
instead aggregate information across frames and spatial regions into
compact memory for future generation.

We present Prediction-Aligned Context Compaction (PACC) to address this question.
The PACC compressor uses the pretrained generator's backbone and is
initialized with its weights. We train it through on-policy distillation  (Figure~\ref{fig:framework}),
using the same frozen generator both as a student when conditioned on
compressed memory and as a teacher when conditioned on the full history.
Following the self-generated rollout strategy of
Self Forcing~\citep{self-forcing}, the student generates continuations
conditioned on memory produced by the current compressor. At each
denoising step, the teacher provides a velocity target for the same
noisy chunk at the same noise level as the student. Both branches also
receive the same text condition and chunks already generated by the
student. We update only the compressor to align the student's velocity
predictions with these targets, teaching it what historical information
to preserve for future generation.

Learning this compaction operation does not require long-video
supervision. We train the compressor using approximately five-second
rollouts, keeping the full-history teacher within the short-horizon regime
where its predictions remain useful. At inference, we apply the compressor
independently to each completed video block, reducing it to one-tenth of
its source tokens, and store the resulting memory in an archive.
Our main inference configuration retrieves a complete compressed block
alongside raw sink and recent frames, maintaining a bounded active context
while the archive grows with video length. Compaction and retrieval thus
play complementary roles: the generator selects among learned summaries
that combine information across frames, rather than relying only on
individual past frames.

We evaluate PACC on MBench~\citep{mbench}, which jointly measures
memory-event coverage and consistency in long-video generation.
In our main comparison, PACC improves the M-score by 6.63 points on Causal-rCM~\citep{causal-rcm}
and 3.19 points on Causal Forcing~\citep{causal-forcing} over the strongest baseline in each panel.
We further evaluate PACC on minute-long MovieGen videos using VBench-Long~\citep{vbench-long}, where it achieves competitive generation quality.
Together, these findings
show that short-rollout supervision can teach useful memory compaction
for long-video generation without modifying the underlying generator.

\section{Related Work}
\label{sec:related-work}

\paragraph{Autoregressive video diffusion.}

Wan~\citep{wan} provides a pretrained video diffusion backbone that jointly denoises video frames using bidirectional attention. Although effective for short clips, extending this formulation to efficient long-video generation requires mechanisms for temporal extrapolation and reuse of historical computation. Recent work adapts such backbones for causal autoregressive generation, denoising successive frames or chunks conditioned on cached history. Self Forcing~\citep{self-forcing} reduces exposure bias by training on self-generated rollouts, while Causal Forcing~\citep{causal-forcing} improves autoregressive distillation using a causally matched teacher. Causal-rCM~\citep{causal-rcm} combines teacher-forcing consistency distillation with self-forcing refinement for few-step streaming video generation and interactive world models.

\paragraph{Memory for long-horizon video generation.}
Existing methods manage historical context through windowed attention
and frame sinks~\citep{yang_2025_longlive}, positional
adaptation~\citep{yesiltepe_2026_infinityrope}, and selective retention or
retrieval of historical frames and KV
chunks~\citep{chen_2026_pyramid,relax-forcing,hu_2026_longliverag,
ye_2026_dysink,yi_2026_worldkv}. Beyond selecting observations,
MemRoPE~\citep{kim_2026_memrope} and Hybrid
Forcing~\citep{li_2026_longhorizon} aggregate evicted history, while
SlotMemory~\citep{dou_2026_slotmemory} organizes KV tokens through
object-centric routing. PackForcing~\citep{pack-forcing} and
Echo-Infinity~\citep{echo-infinity} further learn compressed or evolving
memory jointly with video generation.
However, these approaches do not explicitly require compact memory to
preserve an unchanged generator's response to the original history,
so successful generation need not establish faithful memory replacement.
PACC instead makes the predictive contribution of history the explicit
preservation target, using the frozen generator's full-context
predictions to supervise a compact substitute for the past.
This teaches memory what must survive compression for the generator
that will consume it, rather than relying on generator adaptation to
make a new memory representation usable.

\section{Method}
\label{sec:method}

\begin{figure}[t]
    \centering
    \includegraphics[width=\linewidth]{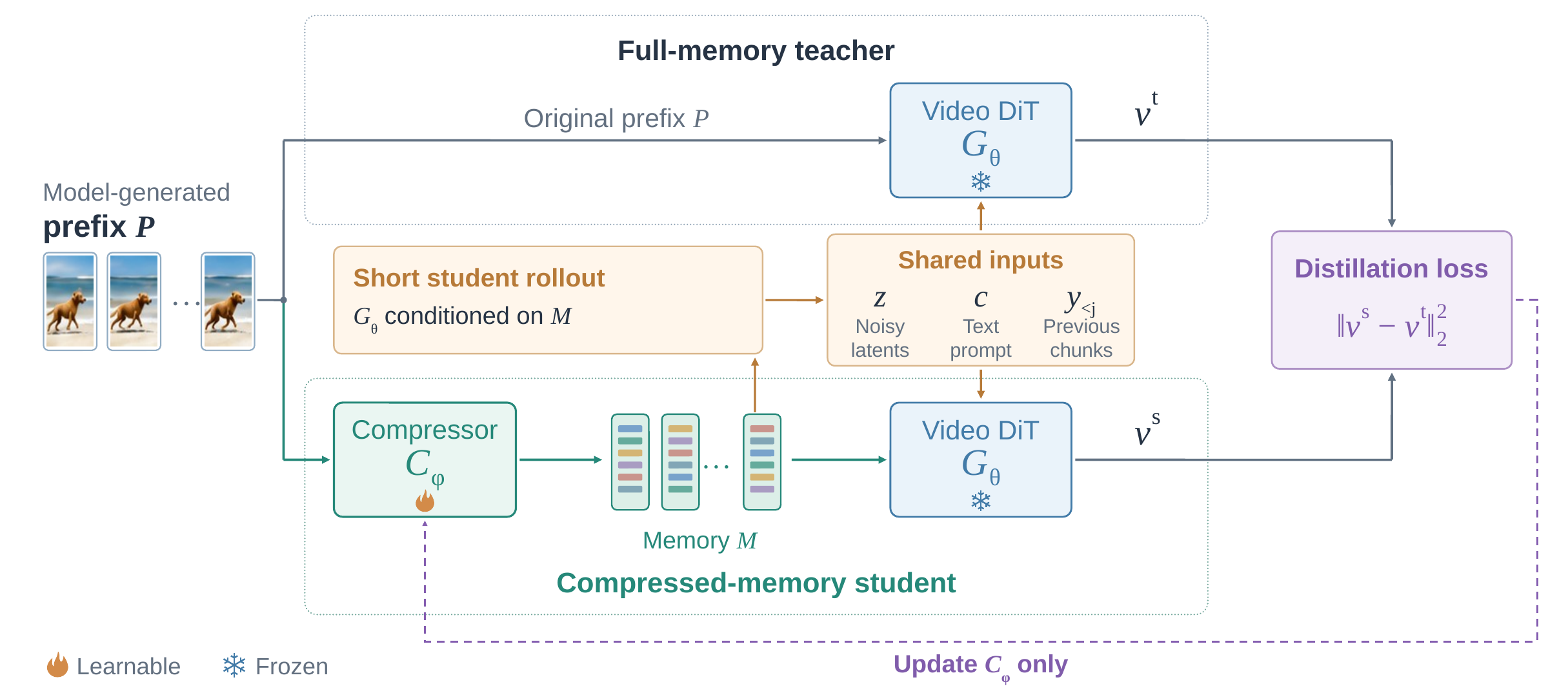}
    \caption{\textbf{PACC training through on-policy distillation.}
    The compressor $C_\phi$ maps a model-generated prefix $P$ to compact
    memory $M=C_\phi(P,c)$. A short student rollout supplies noisy latents
    $z$ and previously generated chunks $y_{<j}$. The full-memory teacher
    and compressed-memory student share the frozen generator $G_\theta$,
    text condition $c$, and rollout inputs, but condition on $P$ and $M$,
    respectively. Matching their velocity predictions $v^{\mathrm{t}}$
    and $v^{\mathrm{s}}$ updates only $C_\phi$ through the student branch.
    The figure abbreviates $z_{j,s}$, $v^{\mathrm{t}}_{j,s}$, and
    $v^{\mathrm{s}}_{j,s}$ as $z$, $v^{\mathrm{t}}$, and $v^{\mathrm{s}}$,
    where $j$ indexes continuation chunks and $s$ indexes denoising steps.
    The shared noise level $t_s$ and the compressor's text-conditioning
    connection are omitted. The loss box shows one squared-error term
    of Equation~\ref{eq:method:loss}.}
    \label{fig:framework}
\end{figure}

Prediction-Aligned Context Compaction (PACC) learns to compress video
history into reusable memory for a frozen autoregressive generator
(Section~\ref{sec:method:compression}). As shown in
Figure~\ref{fig:framework}, we train the compressor through on-policy
distillation. The generator conditioned on the current compressed memory
(student) produces short continuations, while the same generator
conditioned on the full history (teacher) supplies targets for the same
noisy inputs at each denoising step. Only the compressor is updated to
align the student's predictions with these targets
(Section~\ref{sec:method:distillation}).
During long-video generation, we periodically compress completed video
blocks and selectively retrieve their stored memory to condition
subsequent generation (Section~\ref{sec:method:inference}).
\subsection{Compressor Architecture}
\label{sec:method:compression}

We use a Wan-based autoregressive video diffusion generator
$G_\theta$~\citep{wan}, whose parameters $\theta$ remain frozen. The
generator operates on a video autoencoder's compressed representation,
rather than directly on pixels. A \emph{latent frame} is one temporal
slice of this representation; spatial patchification divides each latent
frame into $N$ tokens. All frame counts below refer to latent frames.
The generator produces $h$ frames per autoregressive \emph{chunk},
conditioned on text $c$ and previously generated context. Here, $c$ is
the frozen text-encoder representation of the prompt.

Our compressor $C_\phi$, with trainable parameters $\phi$, converts a
completed segment of video into a shorter sequence of embeddings that
can condition further generation. During training, we call the initial
segment the \emph{prefix} $P$, its compressed representation the
\emph{memory} $M$, and the chunks generated after it the
\emph{continuation}. 


For a prefix of $K$ frames, $C_\phi$ maps its $KN$ spatial tokens to
$Kq$ memory tokens, where $q<N$ is the number of memory queries per
source frame. It embeds the prefix, appends the queries, and retains
only their output embeddings:
\begin{equation}
    M=C_\phi(P,c)
    =\operatorname{LN}\!\left(
        F_\phi\bigl([E_\phi(P);\mathbf{Q}_K],c\bigr)_{\mathrm{mem}}
      \right),
    \qquad M\in\mathbb{R}^{Kq\times d}.
    \label{eq:method:compressor}
\end{equation}
Here, $E_\phi(P)\in\mathbb{R}^{KN\times d}$ embeds the prefix,
$F_\phi$ is the compressor transformer, and $d$ is its hidden width.
The query matrix $\mathbf{Q}_K\in\mathbb{R}^{Kq\times d}$ repeats a
shared learnable $q\times d$ table across the $K$ frames, with positional
coordinates distinguishing the copies. The notation $[\,\cdot\,;\,\cdot\,]$
denotes token concatenation, the subscript $\mathrm{mem}$ selects the
query outputs, and $\operatorname{LN}$ applies layer normalization.
We initialize compatible compressor weights from $G_\theta$ and
initialize the memory queries separately.

The queries attend to the entire prefix, so a memory token can aggregate
information across multiple frames and spatial regions. Prefix tokens
attend to one another; queries attend to both prefix tokens and other
queries, but prefix tokens do not attend back to the queries. We process
the clean prefix without adding noise and fix the compressor's diffusion
noise-level input to $t=0$. Here, $t$ denotes noise level, not video time.
The output $M$ is supplied directly to the generator as memory embeddings,
without decoding it into video frames.

The retained token fraction is $\rho=q/N$.
We place memory tokens consecutively on the generator's spatial positional
grid, using one frame-sized grid per $N$ tokens and retaining any
partially filled final grid. Thus, the memory occupies
$n_M=\lceil Kq/N\rceil$ positional groups instead of the original $K$
frames. For example, 
at $\rho=0.1$ an 18-frame prefix occupies
$\lceil 18\cdot 0.1\rceil=2$ such groups. These groups provide attention positions, not a correspondence
to individual video frames. The compressed-memory branch places its
continuation after these groups, while the full-history branch retains
the original prefix positions. During training, the compressed-memory
branch receives all of $M$ and no additional uncompressed frames from
$P$; it subsequently caches its own generated continuation.

\subsection{On-Policy Distillation from Full-History Predictions}
\label{sec:method:distillation}
\label{sec:method:rollouts}

A useful compact memory should enable the frozen generator to match the
predictions it would make with access to the original prefix. We therefore
compare two evaluations of the same generator: the \emph{compressed-memory
student} conditioned on $M=C_\phi(P,c)$ and the \emph{full-memory teacher}
conditioned on $P$, as shown in Figure~\ref{fig:framework}.
Here, full memory means the original prefix and the previously generated
continuation, not another learned memory representation. Student and
teacher refer to different conditioning, not separate generator parameters.
Their velocity-prediction difference supervises what the compressor
should preserve.

\paragraph{Problem setup.}
Let $J$ be the number of continuation chunks and $S$ the number of
denoising steps per chunk. We index chunks by $j\in\{0,\ldots,J-1\}$
and denoising steps by $s\in\{0,\ldots,S-1\}$. Let $y_j$ be the clean
chunk generated by the student at index $j$, and let
$y_{<j}=[y_0,\ldots,y_{j-1}]$ denote its earlier chunks, with
$y_{<0}=\emptyset$. Both branches receive the same $y_{<j}$, giving
conditioning sequences
$[M,y_{<j}]$ and $[P,y_{<j}]$. To reuse historical computation, each
branch maintains its own key--value (KV) cache, which stores the
transformer's attention keys and values for previously processed context.
The teacher cache $\mathcal{K}^{\mathrm{t}}_j$ represents $P$ and
$y_{<j}$. The student cache $\mathcal{K}^{\mathrm{s}}_j$ represents
$y_{<j}$, while $M$ is supplied separately at each student evaluation.
Although the generated chunks are identical across branches, their cached
features can differ because they were processed under different histories.

Let $D_z$ be the number of scalar latent entries in a chunk. For a clean
chunk $y$ and Gaussian noise $\epsilon\sim\mathcal{N}(0,I_{D_z})$,
where $I_{D_z}$ is the $D_z$-dimensional identity matrix, we use the
rectified-flow convention $z_t=(1-t)y+t\epsilon$, with velocity oriented
as $\epsilon-y$. Chunk tensors are viewed as vectors of length $D_z$
when writing these expressions. Let $\mathbf{t}=(t_0,\ldots,t_S)$,
with $t_0>\cdots>t_S=0$, be the generator's $S$-step sampling schedule.
For continuation chunk $j$, let $z_{j,s}$ be the student's noisy input
at denoising step $s$ and noise level $t_s$.

\paragraph{Prediction-matching loss.}
The paired predictions are
\begin{align}
    v^{\mathrm{s}}_{j,s}
    &=G_\theta\!\left(z_{j,s},t_s,c;
          \mathcal{K}^{\mathrm{s}}_j,M\right),
          \label{eq:method:student}\\
    v^{\mathrm{t}}_{j,s}
    &=G_\theta\!\left(z_{j,s},t_s,c;
          \mathcal{K}^{\mathrm{t}}_j\right).
          \label{eq:method:teacher}
\end{align}
The superscripts $\mathrm{s}$ and $\mathrm{t}$ are student and teacher
labels, distinct from the step index $s$ and noise level $t$.
Figure~\ref{fig:framework} suppresses the $(j,s)$ subscripts on the
noisy input and velocity predictions. Both branches receive the same
noisy chunk, noise level, text condition, and previously generated
continuation. The text condition is supplied with the example; it is
not generated by the student rollout. The teacher supplies a prediction
target at the student's current state, not a separate target video.

The distillation objective averages these prediction differences over
all continuation chunks and denoising steps:
\begin{equation}
    \mathcal{L}_{\mathrm{dist}}
    =\frac{\lambda}{JS}
      \sum_{j=0}^{J-1}\sum_{s=0}^{S-1}
      \frac{1}{D_z}
      \left\|v^{\mathrm{s}}_{j,s}
      -v^{\mathrm{t}}_{j,s}\right\|_2^2,
    \label{eq:method:loss}
\end{equation}
where $\lambda$ is a constant loss multiplier. Teacher predictions are
computed without gradient tracking, and only $C_\phi$ is optimized.
The normalization by $JS$ gives each example equal total weight
regardless of its continuation length; example losses are then averaged
across the minibatch. Matching predictions over multiple future chunks
encourages memory to remain useful beyond the immediate next prediction.

\paragraph{On-policy continuations.}
Compression changes the generator's conditioning and can therefore change
the continuation it produces. Training only on fixed continuations may
miss the states encountered when generation actually depends on the
learned memory. Following the self-generated-context principle of
Self Forcing~\citep{self-forcing}, we generate each continuation with
the current compressor and frozen generator. The teacher evaluates the
states visited by this student rollout, including those affected by
earlier compression errors. Training is therefore on-policy with respect
to the current compressed-memory system.

The initial prefix and the continuation have different sources.
We sample a pair $(x,c)$ from a fixed dataset
$\mathcal{D}_{\mathrm{roll}}$ of short videos generated by the same
backbone, where $x$ is the clean latent video and $c$ its text condition.
We use half-open indexing, so $x_{0:K}$ denotes its first $K$ frames.
After taking the first $K$ frames as $P$, we generate a new continuation
using the current $C_\phi$. Stored frames after $P$ are neither
prediction targets nor inputs to either continuation cache.

\paragraph{Why short rollouts?}
The full-history generator is a useful teacher only where its own
predictions remain reliable. In our setting, extending its autoregressive
rollouts far beyond the short-video training regime produces substantial
quality degradation, even when the generated history is retained in full.
Related work also reports long-rollout drift and limitations of
short-clip teachers for supervising long
sequences~\citep{yang_2025_longlive,relax-forcing}.
We therefore obtain distillation targets within approximately five-second
rollouts rather than imitate degraded long-video behavior. Short training
is a choice about supervision quality, not only computational cost.
The trained compressor is later
applied repeatedly to completed blocks during longer generation.

\paragraph{Sampling and gradient flow.}
Using the sampling schedule $\mathbf{t}$, each chunk starts from
$z_{j,0}=t_0\epsilon_j$, where
$\epsilon_j\sim\mathcal{N}(0,I_{D_z})$. After computing the paired
predictions and accumulating Equation~\ref{eq:method:loss}, we use
the student's velocity to estimate a clean chunk and then re-noise it
at the next noise level:
\begin{align}
    \widehat{y}_{j,s}
    &=\operatorname{sg}\!\left(
          z_{j,s}-t_s v^{\mathrm{s}}_{j,s}\right),
          \label{eq:method:clean-estimate}\\
    z_{j,s+1}
    &=\operatorname{sg}\!\left(
          (1-t_{s+1})\widehat{y}_{j,s}
          +t_{s+1}\epsilon_{j,s}\right),
    \quad \epsilon_{j,s}\sim\mathcal{N}(0,I_{D_z}),
          \label{eq:method:renoise}
\end{align}
where $\operatorname{sg}$ denotes stop-gradient and
$\widehat{y}_{j,s}$ is the current clean-chunk estimate. Noise is sampled
independently across frames and sampling transitions; the resulting
$z_{j,s}$ is shared by both branches. At $t_S=0$, the completed chunk
is $y_j=z_{j,S}$.

Gradients from each prediction-matching term pass through the frozen
student generator into $M$ and then $C_\phi$. Freezing $\theta$ does
not block gradients with respect to the generator's memory input.
Teacher evaluations, sampling transitions, and cache updates are performed
without propagating gradients through them. Thus, one memory receives
supervision from multiple future chunks, but we do not backpropagate
through earlier denoising steps or generated chunks.
When appending a completed chunk to the student cache, the memory
input is detached, so the cache itself does not carry gradients from $M$.

\begin{algorithm}[t]
\caption{Training PACC through on-policy distillation}
\label{alg:method:train}
\small
\begin{algorithmic}[1]
\Input Frozen generator $G_\theta$; trainable compressor $C_\phi$;
    dataset $\mathcal{D}_{\mathrm{roll}}$ of $T$-frame latent-video/text pairs $(x,c)$
\Input Total horizon $T$; chunk size $h$ with $h\mid T$ and $T>h$;
    chunk dimension $D_z$
\Input $S$-step schedule $\mathbf{t}=(t_0,\ldots,t_S)$,
    $t_0>\cdots>t_S=0$; loss multiplier $\lambda$; optimizer $\mathcal{O}$
\Output Updated compressor parameters $\phi$; generator parameters $\theta$ unchanged
\State Sample $(x,c)\sim\mathcal{D}_{\mathrm{roll}}$
\State Sample $K\sim\operatorname{Uniform}\{h,2h,\ldots,T-h\}$
\State $P\gets x_{0:K}$; $J\gets(T-K)/h$
    \Comment{$x_{0:K}$ contains the first $K$ frames}
\State $M\gets C_\phi(P,c)$ \Comment{Retain the compressor graph}
\State $\mathcal{K}^{\mathrm{t}}_0\gets\operatorname{Prefill}_\theta(P,c)$;
    $\mathcal{K}^{\mathrm{s}}_0\gets\emptyset$; $\mathcal{L}_{\mathrm{dist}}\gets0$
    \Comment{Prefill: process clean prefix at $t{=}0$, no grad}
\For{$j=0,\ldots,J-1$}
    \State Draw $\epsilon_j\sim\mathcal{N}(0,I_{D_z})$;
        $z_{j,0}\gets t_0\epsilon_j$
    \For{$s=0,\ldots,S-1$}
        \State $v^{\mathrm{s}}_{j,s}\gets G_\theta(z_{j,s},t_s,c;
            \mathcal{K}^{\mathrm{s}}_j,M)$ \Comment{Gradients to $M$}
        \State $v^{\mathrm{t}}_{j,s}\gets G_\theta(z_{j,s},t_s,c;
            \mathcal{K}^{\mathrm{t}}_j)$ \Comment{No gradient tracking}
        \State $\mathcal{L}_{\mathrm{dist}}\gets\mathcal{L}_{\mathrm{dist}}
            +\frac{\lambda}{JSD_z}\|v^{\mathrm{s}}_{j,s}-v^{\mathrm{t}}_{j,s}\|_2^2$
        \State $\widehat y_{j,s}\gets\operatorname{sg}
            (z_{j,s}-t_s v^{\mathrm{s}}_{j,s})$
        \State Draw $\epsilon_{j,s}\sim\mathcal{N}(0,I_{D_z})$
        \State $z_{j,s+1}\gets\operatorname{sg}
            ((1-t_{s+1})\widehat y_{j,s}+t_{s+1}\epsilon_{j,s})$
    \EndFor
    \State $y_j\gets z_{j,S}$
    \State $\mathcal{K}^{\mathrm{t}}_{j+1}\gets
        \operatorname{Append}_\theta(\mathcal{K}^{\mathrm{t}}_j,y_j,c)$
        \Comment{Append: add chunk KV at $t{=}0$, no grad}
    \State $\mathcal{K}^{\mathrm{s}}_{j+1}\gets
        \operatorname{Append}_\theta(\mathcal{K}^{\mathrm{s}}_j,y_j,c;
        \operatorname{sg}(M))$
\EndFor
\State $\phi\gets\operatorname{Step}_{\mathcal{O}}
    (\phi,\nabla_\phi\mathcal{L}_{\mathrm{dist}})$
\end{algorithmic}
\end{algorithm}

\subsection{Long-Video Inference}
\label{sec:method:inference}

At inference, both $G_\theta$ and the trained $C_\phi$ remain frozen.
We generate video in native chunks of $h$ latent frames and periodically
compress each completed block of $L$ frames, where $h$ divides $L$.
For a block $X_b$ generated with text condition $c_b$, we compute
$M_b=C_\phi(X_b,c_b)$ and append it to a memory archive. Each block is
compressed once; earlier memories are stored without recompression.

\paragraph{Memory with sink and recent frames.}
Following prior work~\citep{yang_2025_longlive,relax-forcing}, we retain
two initial uncompressed latent frames as sink anchors and the latest
frame for recent context. Rather than read the growing archive in full,
we select one complete compressed block at each chunk boundary using a
fixed key-similarity rule. Its memory tokens are placed between the sink
and recent frames and remain active throughout the next chunk's denoising
steps. Thus, compression expands the archive every $L$ frames, while
retrieval refreshes the selected memory every $h$ frames. For a fixed
block length and compression rate, this keeps the active context bounded
even as the stored archive grows. Compaction and retrieval are therefore
complementary: the generator selects among learned summaries of whole
blocks rather than relying only on individual retained frames. We provide
the selection score, cache construction, positional assignment, and
inference pseudocode in Appendix~\ref{app:pacc-retrieval}.

\section{Experiments}
\label{sec:experiments}

We evaluate PACC's long-horizon memory preservation and generation quality
on two frozen video backbones. We compare against alternative
context-management policies (Section~\ref{sec:exp_comparison}) and
examine key design choices through ablations (Section~\ref{sec:ablation}).

\begin{table}[t]
\centering
\caption{\textbf{Long-horizon memory and generation quality.}
MBench evaluates 26-second videos; VBench-Long evaluates minute-long
videos on MovieGenBench.
Human, Object, and Causal average the paired M-scores for
identity/appearance, geometry/texture, and state/correctness, respectively.
M-score averages all six memory dimensions; VBench-Long Avg.\ averages
six quality dimensions and is not the official VBench total.
Scores are on a 0--100 scale; higher is better. Bold and underlining
mark the best and second-best values in each column within each backbone.}
\label{tab:exp_main_comparison}

\setlength{\tabcolsep}{4pt}
\renewcommand{\arraystretch}{1.08}
\resizebox{0.9\linewidth}{!}{%
\begin{tabular*}{\linewidth}{@{\extracolsep{\fill}}lcccc@{\hspace{1.5em}}c@{}}
\toprule
\multirow{2}{*}{Method}
    & \multicolumn{4}{c}{MBench $\uparrow$}
    & VBench-Long $\uparrow$ \\
\cmidrule(lr){2-5}\cmidrule(l){6-6}
    & Human & Object & Causal & M-score & Avg. \\
\midrule

\multicolumn{6}{@{}l}{\textit{Causal-rCM c3-3}} \\
\addlinespace[2pt]
Infinity-RoPE~\citep{yesiltepe_2026_infinityrope}
    & 31.83 & 23.83 & 60.92 & 38.86 & 80.09 \\
Relax Forcing~\citep{relax-forcing}
    & 27.10 & \underline{36.21} & \underline{65.74}
    & \underline{43.01} & \textbf{81.00} \\
Rolling Sink~\citep{li_2026_rolling}
    & 15.96 & 4.76 & 42.25 & 20.99 & 77.43 \\
Deep Forcing~\citep{yi_2025_deep}
    & 31.57 & 34.56 & 49.20 & 38.44 & 79.94 \\
MemRoPE~\citep{kim_2026_memrope}
    & \underline{31.97} & 31.01 & 60.27 & 41.08 & 79.64 \\
\addlinespace[2pt]
\textbf{PACC (Ours)}
    & \textbf{40.60} & \textbf{41.00} & \textbf{67.32}
    & \textbf{49.64} & \underline{80.50} \\

\midrule
\multicolumn{6}{@{}l}{\textit{Causal Forcing}} \\
\addlinespace[2pt]
Infinity-RoPE~\citep{yesiltepe_2026_infinityrope}
    & 49.92 & \underline{52.82} & 65.28 & 56.00 & 83.04 \\
Relax Forcing~\citep{relax-forcing}
    & \underline{51.88} & 50.71 & \underline{73.17}
    & \underline{58.58} & \textbf{83.79} \\
Rolling Sink~\citep{li_2026_rolling}
    & 50.52 & 44.90 & 62.06 & 52.49 & 82.16 \\
Deep Forcing~\citep{yi_2025_deep}
    & 45.39 & 51.69 & 62.42 & 53.17 & 83.14 \\
MemRoPE~\citep{kim_2026_memrope}
    & 44.78 & 52.16 & 67.17 & 54.70 & 83.55 \\
\addlinespace[2pt]
\textbf{PACC (Ours)}
    & \textbf{55.70} & \textbf{54.15} & \textbf{75.46}
    & \textbf{61.77} & \underline{83.57} \\

\bottomrule
\end{tabular*}
}
\end{table}

\subsection{Experimental Setup}
\label{sec:exp_setup}

\paragraph{Backbone and optimization.}
Our main experiments use the four-step, 1.3B Causal-rCM c3-3
model~\citep{causal-rcm}, and Causal Forcing model~\citep{causal-forcing}, both built on Wan~\citep{wan}, with $h=3$ latent
frames per autoregressive chunk. We train only the compressor $C_\phi$;
the generator $G_\theta$, video autoencoder, and text encoder remain
frozen. We use 81,920 model-generated rollout records for training and
256 held-out records (approximately five seconds each). Training uses a global batch size of eight and AdamW
with a peak learning rate of $10^{-4}$ and a staged cosine schedule. 


\paragraph{PACC configurations.}
Each training example spans a $T=21$-frame horizon.
We sample the prefix length $K$ uniformly from $\{3,6,9,12,15,18\}$ and
generate the remaining $J=(T-K)/h$ continuation chunks on-policy with the
current compressor, so the memory must support between one and six future
chunks. The distillation objective of
Equation~\ref{eq:method:loss} is applied at all $S=4$ denoising steps of
every continuation chunk, with loss multiplier $\lambda=100$.
Appendix~\ref{app:exp_protocol} gives the remaining optimization settings.
By default, we use the 5,000-update checkpoint with compression factor
$\rho^{-1}=10$ and whole-block retrieval
(Section~\ref{sec:method:inference}): one complete compressed block is
selected alongside two sink frames and the latest frame, keeping active
context bounded. 

\paragraph{Benchmarks and baselines.}
We evaluate memory preservation with MBench~\citep{mbench}, reporting
six human, object, and causal consistency dimensions on approximately
26-second videos. The aggregate M-score averages the dimension-level
scores, each combining memory-event coverage and conditional reliability.
For longer generation, we evaluate approximately one-minute videos on
the first 128 MovieGenBench prompts~\citep{moviegen}, with five seeds per
prompt, using VBench-Long~\citep{vbench,vbench-long}. We report six
visual-quality dimensions.
Benchmark composition, metric definitions, and evaluation procedures are
given in Appendix~\ref{app:benchmark-protocols}.
We compare inference-time policies on the same frozen generator within
each backbone: Infinity-RoPE~\citep{yesiltepe_2026_infinityrope},
Relax Forcing~\citep{relax-forcing},
Rolling Sink~\citep{li_2026_rolling},
Deep Forcing~\citep{yi_2025_deep}, and
MemRoPE~\citep{kim_2026_memrope}. Policy settings are provided
in Appendix~\ref{app:baselines}.


\subsection{Comparison with Baselines}
\label{sec:exp_comparison}

\paragraph{Long-horizon memory.}
Table~\ref{tab:exp_main_comparison} summarizes memory preservation and
minute-long generation quality on both backbones. PACC achieves M-scores
of 49.64 on Causal-rCM and 61.77 on Causal Forcing, exceeding the strongest
baseline in this comparison by 6.63 and 3.19 points, respectively.
It also leads all three aggregated memory categories---human, object,
and causal consistency---on both backbones.

\begin{figure}[t]
    \centering
    \includegraphics[width=\linewidth]{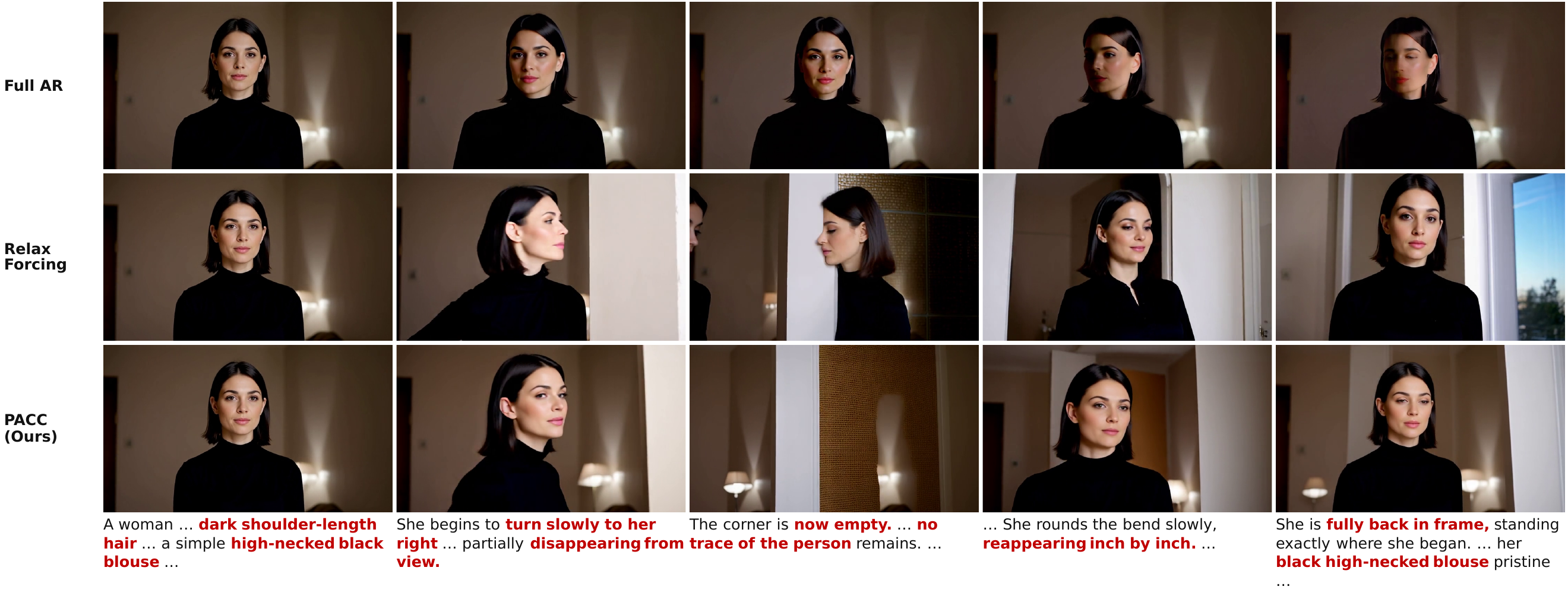}
    
    \caption{\textbf{Qualitative memory comparison on Causal Forcing.}
    Columns follow five consecutive prompt segments, with excerpts shown
    below. The subject is asked to leave the frame and later return with
    her appearance unchanged. PACC depicts the empty-scene stage and
    subsequent return while retaining her hairstyle and black
    high-necked blouse. In contrast, the subject remains visible in both
    baselines at the displayed empty-scene stage.}
    \label{fig:qualitative_memory}

\end{figure}

\begin{figure}[t]

    \centering
    \captionsetup[subfigure]{font=footnotesize,labelfont=bf,
        justification=centering,singlelinecheck=false,skip=2pt}
    {\footnotesize\bfseries
    \makebox[0.49\linewidth][c]{Training duration ($\rho^{-1}=10$)}\hfill
    \makebox[0.49\linewidth][c]{Compression factor (2k updates)}\par}
    \vspace{2pt}
    \begin{subfigure}[t]{0.238\linewidth}
        \centering
        \includegraphics[width=\linewidth]{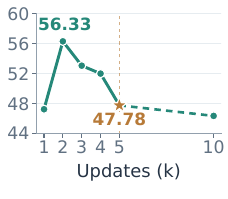}
        \caption{M-score $\uparrow$}
        \label{fig:ablation_updates_memory}
    \end{subfigure}\hfill
    \begin{subfigure}[t]{0.238\linewidth}
        \centering
        \includegraphics[width=\linewidth]{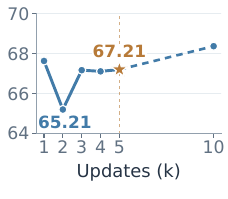}
        \caption{Imaging quality $\uparrow$}
        \label{fig:ablation_updates_imaging}
    \end{subfigure}\hfill
    \begin{subfigure}[t]{0.238\linewidth}
        \centering
        \includegraphics[width=\linewidth]{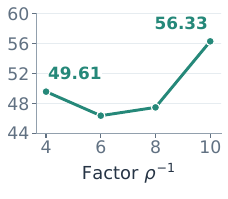}
        \caption{M-score $\uparrow$}
        \label{fig:ablation_compression_memory}
    \end{subfigure}\hfill
    \begin{subfigure}[t]{0.238\linewidth}
        \centering
        \includegraphics[width=\linewidth]{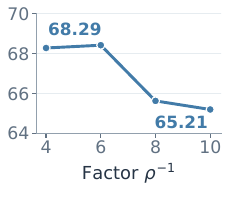}
        \caption{Imaging quality $\uparrow$}
        \label{fig:ablation_compression_imaging}
    \end{subfigure}
    \caption{\textbf{Training and compression affect memory and imaging
    quality differently.} (a,b) Training-duration sweep at $\rho^{-1}=10$;
    stars mark the 5,000-update main checkpoint. (c,d) Compression-factor
    sweep at 2,000 updates. M-score uses the MBench subset; imaging quality
    uses 64 minute-long MovieGen videos per setting. The complete sweep and
    complementary quality dimensions are reported in
    Appendix~\ref{app:ablation_details}. Lines are visual guides.}
    \label{fig:ablation_tradeoffs}
\end{figure}

Figure~\ref{fig:qualitative_memory} provides a qualitative example on
Causal Forcing: PACC depicts the prompted departure-and-return sequence
while preserving the subject's appearance in the displayed frames.
An additional occlusion-and-reveal example on Causal-rCM is provided in
Appendix~\ref{app:qualitative_results}
(Figure~\ref{fig:qualitative_memory_rcm}).

\paragraph{Minute-long generation quality.}
PACC maintains competitive quality, with VBench-Long averages of 80.50
on Causal-rCM and 83.57 on Causal Forcing. These rank second within each
backbone, only 0.50 and 0.22 points below the best baseline, respectively. Full per-dimension memory and quality
results are reported in Appendix~\ref{app:detailed_results}.

Together, these results show improved long-horizon memory preservation
with competitive minute-long generation quality on both frozen backbones.

\subsection{Ablation Study}
\label{sec:ablation}

We examine training duration and compression factor on MBench and
VBench-Long subsets on Causal-rCM. Full per-dimension results are provided in
Appendix~\ref{app:ablation_details}.

\paragraph{Training duration.}
Figure~\ref{fig:ablation_tradeoffs}(a,b) shows that memory preservation
and imaging quality favor different checkpoints. At $\rho^{-1}=10$,
memory performance peaks at 2,000 updates with an M-score of 56.33.
The 5,000-update checkpoint used in our main experiments has higher
imaging quality (67.21 versus 65.21) and aesthetic quality
(61.14 versus 60.09), but lower memory performance and dynamic degree.
Relative to the early memory peak, our main checkpoint thus favors
imaging and aesthetic quality. The full sweep is non-monotonic and
does not identify a single checkpoint that maximizes both memory
and quality.

\paragraph{Compression factor.}
Figure~\ref{fig:ablation_tradeoffs}(c,d) compares compression factors
at 2,000 updates. Tenfold compression achieves the highest M-score
in this sweep, improving over fourfold compression from 49.61 to 56.33
while retaining 60\% fewer memory tokens per source block. Its imaging
quality is lower, however (65.21 versus 68.29). Retaining more tokens
therefore does not necessarily improve deployed memory performance
at this training duration. This rate sweep is separate from the
5,000-update main configuration.

\section{Conclusion}

We presented Prediction-Aligned Context Compaction (PACC), which learns
to compress video history into reusable memory for a frozen autoregressive
generator. Through on-policy distillation, we train only the compressor
to produce memory that enables the generator to approximate its predictions
under full-history conditioning. The compressor learns from short rollouts
and is applied repeatedly during long-video generation. At inference,
completed video blocks form a compressed memory archive, from which
selective retrieval maintains a bounded active context. Compaction thus
complements selection: rather than relying only on individual retained
frames, the generator can use learned representations that aggregate
information across the history. Our MBench and Vbench-Long experiments on Causal-rCM and Causal Forcing demonstrate
the benefits of learned compaction for long-horizon memory preservation without hurting the generation quality.



\subsubsection*{Acknowledgments}
LEK and WG have been supported in part by NSF 2336612 and Rice University Funds.

\bibliography{ref}
\bibliographystyle{iclr2027_conference}
\newpage
\appendix

\section{Long-Video Inference Details}
\label{app:pacc-retrieval}

We detail the periodic compaction and whole-block retrieval procedure in
Section~\ref{sec:method:inference}. The generator $G_\theta$ and trained
compressor $C_\phi$ remain frozen throughout inference. Retrieval uses a
fixed rule, and all operations run without gradient tracking.

\paragraph{Periodic compaction and memory storage.}
We first describe generation of $B$ complete blocks, each containing
$L$ latent frames and $L/h$ native chunks of $h$ frames, with $h\mid L$.
Let $X_b$ denote block $b\in\{0,\ldots,B-1\}$ and $c_b$ its text
condition. Each source frame contains $N$ spatial tokens and contributes
$q$ memory queries. Thus, a completed block produces $Q=Lq$ memory tokens,
with retained fraction $\rho=q/N$ and compression factor $\rho^{-1}$.
We compress the block and append its memory to the archive:
\begin{equation}
    M_b=C_\phi(X_b,c_b),
    \qquad
    \mathcal{A}_{b+1}=\mathcal{A}_b\mathbin{\|}(M_b,\Pi_b),
    \label{eq:method:archive}
\end{equation}
where $\mathcal{A}_0=\emptyset$, $\|$ denotes sequence append, and
$\Pi_b$ contains the source-block-aligned coordinates of the packed memory
tokens. Each block is compressed once, without recompressing earlier
memories. We retain all output tokens, including a partially filled final
positional group. The default uses $L=21$, $h=3$, and $\rho^{-1}=10$.
Compression of the final block is unnecessary when no generation follows.


\paragraph{Incremental memory KV construction.}
Before generating $X_b$, we save its active past key--value (KV) features
and attention positions as $\mathcal{H}_b$, with
$\mathcal{H}_0=\emptyset$. After compression, a single frozen-generator
pass constructs the new memory's layer-wise KV features:
\begin{equation}
    F_b=\operatorname{KVNew}_\theta
        (M_b,c_b,\Pi_b;\mathcal{H}_b,t=0).
    \label{eq:method:incremental-kv}
\end{equation}
Here, $t=0$ denotes zero diffusion noise. Memory tokens attend
bidirectionally to one another and read the saved past KV without
recomputing past queries. We store $F_b$ in a KV archive $\mathcal{B}$,
with $\mathcal{B}[b]=F_b$. Keys are stored before rotary positional
encoding (RoPE); we refer to them as \emph{canonical keys}. Previously
stored entries remain unchanged, so later prompts and retrieval decisions
do not require jointly reprocessing the archive.

\paragraph{Candidate construction.}
Let $r$ be the number of latent frames already generated. Once memory is
available, we refresh retrieval at each native chunk boundary, every
$h$ frames. We adapt the Relax Forcing candidate
construction~\citep{relax-forcing}: its grid is applied to the
chronological list of complete $N$-token memory units, and shortlisted
units are mapped to distinct source blocks. The resulting eligible set
$\mathcal{E}_r$ spans the stored history without a later-half restriction,
but need not contain every archived block. Residual tokens are not
separate shortlist entries; they remain part of their source block when
it is scored or retrieved.


\paragraph{Selection score.}
Selection uses canonical keys from the first transformer layer. Let
$p_i$, $p_{\mathrm{sink}}$, and $p_{\mathrm{recent},r}$ denote normalized
key prototypes for candidate block $i$, the two sink frames together,
and the latest raw frame, respectively. Here, \emph{raw} means
uncompressed latent-frame features, not pixels. Each prototype averages
keys over tokens and then attention heads, followed by normalization.
Following the Relax Forcing scoring rule~\citep{relax-forcing},
\begin{equation}
\begin{aligned}
    \sigma_{i,r}
      &=\langle p_i,p_{\mathrm{sink}}\rangle
        -2\langle p_i,p_{\mathrm{recent},r}\rangle,\\
    i_r&=\operatorname*{arg\,max}_{i\in\mathcal{E}_r}\sigma_{i,r}.
\end{aligned}
\label{eq:method:block-selection}
\end{equation}
The score favors similarity to the sink anchors and discourages redundancy
with recent context. All layers use the same selected block index $i_r$,
but each reads its own KV features for all $Q$ tokens of that block.
The $N$-token units used to form candidates are therefore not the
retrieval units of the final context.

\paragraph{Active context and cache updates.}
At transformer layer $\ell$, the active past context is
\begin{equation}
    \mathcal{K}^{(\ell)}_r
        =[S_0^{(\ell)},S_1^{(\ell)},F_{i_r}^{(\ell)},R_{r-1}^{(\ell)}],
    \label{eq:method:hybrid}
\end{equation}
where $S_0^{(\ell)}$ and $S_1^{(\ell)}$ are the first two raw frames'
KV pairs, $F_{i_r}^{(\ell)}$ is the selected memory block's KV, and
$R_{r-1}^{(\ell)}$ is the latest raw frame's KV. Brackets denote token
concatenation. We use zero-based indexing, so the latest frame is $r-1$.

The past context remains fixed during the chunk's denoising steps. A
subsequent clean-cache pass at $t=0$ updates the recent-frame features.
Only the two sinks and the latest frame persist as raw anchors; the
remaining frames of the unfinished block are retained as latents for
compression, without accumulating their raw KV in active attention.
Before the first compressed block is available, generation uses the
raw-frame selection policy with the context available at each step.

\paragraph{Positional assignment and context budget.}
At attention time, canonical keys receive compact temporal positions in
the order of two sink slots, $n_{\mathrm{slot}}=\lceil Q/N\rceil$
memory slots, and one recent-frame slot, immediately before the query
chunk. The last memory slot may be partially filled. Explicit spatial
coordinates preserve the recent frame's full grid without padding or
discarding memory tokens. Assigning these positions does not modify the
stored canonical KV.

Once memory is available, active past context contains $3N+Q$ tokens
per layer, excluding the query chunk. With $L=21$, $N=1{,}560$, and
$\rho=0.1$,
\begin{equation}
    Q=Lq=\rho LN=3{,}276,
    \qquad 3N+Q=7{,}956.
    \label{eq:exp:block-token-count}
\end{equation}
The corresponding raw-frame policy uses $4N=6{,}240$ tokens for two
sinks, one selected historical frame, and one recent frame. Whole-block
retrieval therefore keeps active context bounded but is not token-budget
matched to that policy. The embedding and KV archives still grow with
video length; bounded attention context does not imply bounded storage.

\paragraph{Inference pseudocode.}
Algorithm~\ref{alg:method:inference} gives the complete-block procedure.
The raw cache $\mathcal{R}$ stores the sink and recent-frame features,
plus candidate raw history during the first block. We write $y_{b,j}$
for chunk $j$ of block $b$, $\mathcal{V}$ for selected historical KV,
and $\mathcal{K}_r$ for active context across all layers.
$\operatorname{SelectFrame}$ returns raw historical KV during bootstrap,
or an empty unit when none is available. $\operatorname{SelectBlock}$
uses the candidate construction and Equation~\ref{eq:method:block-selection}.
$\operatorname{Compose}$ combines selected history with available raw
anchors and assigns attention positions; $\operatorname{Snapshot}$
saves the resulting past KV and positions.

$\operatorname{SampleChunk}_\theta$ uses the sampling updates in
Section~\ref{sec:method:rollouts}, with chunk dimension $D_z$ and noise
schedule $\mathbf{t}=(t_0,\ldots,t_S)$. The clean-cache operation
$\operatorname{UpdateRaw}_\theta$ retains bootstrap candidates in block
$b=0$, and only sink and recent-frame features thereafter.
$\operatorname{KeepAnchors}$ removes the remaining bootstrap candidates
once compressed memory is available.

\begin{algorithm}[t]
\caption{Periodic compaction and whole-block retrieval}
\label{alg:method:inference}
\small
\begin{algorithmic}[1]
\Input Frozen $G_\theta$ and $C_\phi$; positive block count $B$;
    text conditions $\{c_b\}_{b=0}^{B-1}$
\Input Block length $L$; chunk size $h$ with $h\mid L$;
    $N$ spatial tokens per frame; chunk dimension $D_z$
\Input Noise schedule $\mathbf{t}=(t_0,\ldots,t_S)$,
    $t_0>\cdots>t_S=0$
\Input Packed memory coordinates $\{\Pi_b\}_{b=0}^{B-2}$;
    selectors $\operatorname{SelectFrame}$ and $\operatorname{SelectBlock}$
\Output Latent video $X$ of $BL$ frames; no parameter updates
\State $\mathcal{A},\mathcal{B},\mathcal{R},X\gets\emptyset$
\For{$b=0,\ldots,B-1$}
    \State $X_b\gets\emptyset$
    \For{$j=0,\ldots,L/h-1$}
        \State $r\gets bL+jh$
        \If{$\mathcal{B}=\emptyset$}
            \State $\mathcal{V}\gets\operatorname{SelectFrame}(\mathcal{R},r;N)$
            \Comment{Raw-history bootstrap}
        \Else
            \State $i_r\gets\operatorname{SelectBlock}(\mathcal{B},\mathcal{R},r;N)$
            \State $\mathcal{V}\gets\mathcal{B}[i_r]$
        \EndIf
        \State $\mathcal{K}_r\gets\operatorname{Compose}(\mathcal{R},\mathcal{V},r;N)$
        \If{$j=0$ and $b<B-1$}
            \State $\mathcal{H}_b\gets\operatorname{Snapshot}(\mathcal{K}_r)$
        \EndIf
        \State $y_{b,j}\gets\operatorname{SampleChunk}_\theta
            (c_b;\mathcal{K}_r,h,D_z,\mathbf{t})$
        \State $\mathcal{R}\gets\operatorname{UpdateRaw}_\theta
            (\mathcal{R},\mathcal{K}_r,y_{b,j},c_b;b)$
        \State $X_b\gets X_b\mathbin{\|}y_{b,j}$
    \EndFor
    \State $X\gets X\mathbin{\|}X_b$
    \If{$b<B-1$}
        \State $M_b\gets C_\phi(X_b,c_b)$
        \State $F_b\gets\operatorname{KVNew}_\theta
            (M_b,c_b,\Pi_b;\mathcal{H}_b,t=0)$
        \State $\mathcal{A}\gets\mathcal{A}\mathbin{\|}(M_b,\Pi_b)$
        \State $\mathcal{B}\gets\mathcal{B}\mathbin{\|}F_b$
        \State $\mathcal{R}\gets\operatorname{KeepAnchors}(\mathcal{R})$
    \EndIf
\EndFor
\State \Return $X$
\end{algorithmic}
\end{algorithm}

\section{Additional Experimental Details and Results}
\label{app:exp_protocol}

\subsection{Training Protocol}

\paragraph{Backbone and compressor.}
The primary generator is the released Causal-rCM Wan2.1 T2V 1.3B c3-3
four-step checkpoint. Its compressor has 30 transformer blocks of width
$d=1{,}536$ and processes the entire sampled prefix. Compatible weights
are initialized from the generator; memory queries are initialized
separately. At $832\times480$ resolution, each latent frame contains
$N=1{,}560$ spatial tokens. Compression factors $\rho^{-1}\in\{4,6, 8,10\}$
correspond to $q\in\{390,260, 195,156\}$ queries per source frame, respectively.
The generator, video autoencoder, and text encoder remain frozen.

\paragraph{Training examples and supervision.}
We use 81,920 model-generated rollout records and 256 held-out records.
Each training example has total horizon $T=21$ latent frames. We sample
prefix length $K$ uniformly from $\{3,6,9,12,15,18\}$ and generate the
remaining continuation online in chunks of $h=3$ frames. The student
represents the initial prefix only through compressed memory and caches
its own completed continuation chunks. At all $S=4$ denoising steps,
the full-history branch supplies conditional velocity targets at the
student's current state. Prediction is conditional-only, with effective
guidance scale one. Gradients pass through the frozen student generator
to $C_\phi$; teacher evaluations, sampling transitions, and appended KV
features do not propagate gradients.

\paragraph{Optimization.}
Training uses a global batch size of eight and AdamW with
$(\beta_1,\beta_2)=(0.9,0.999)$, numerical stabilizer $10^{-8}$,
weight decay $0.01$, and global gradient-norm clipping at $5$.
After 100 warmup updates, a continuous staged cosine schedule decays the
learning rate from $10^{-4}$ to $5\times10^{-5}$ at 2,500 updates,
$2\times10^{-5}$ at 5,000 updates, and $5\times10^{-6}$ at 10,000
updates. At this batch size, 5,000 updates correspond to 40,000 rollout
record exposures, not a full pass through the training corpus.

\subsection{Benchmark Protocols}
\label{app:benchmark-protocols}

\paragraph{MBench composition and generation.}
MBench~\citep{mbench} organizes memory capability into entity,
environment, and causal consistency, with twelve sub-dimensions.
Its text-conditioned setting, MBench-T, includes eleven sub-dimensions
and 684 cases: 120 human, 100 object, 391 causal, and 73 environment.
Following the benchmark's human-correlation
study~\citep[Table~3]{mbench}, we report six dimensions with Spearman
correlations of 0.69--0.97: human identity and appearance, object geometry
and texture, and causal state progress and progress correctness. The
last corresponds to physical plausibility in the benchmark terminology.
These dimensions cover a nominal population of 611 cases before evaluator
exclusions; the environment subset is not included.

Each case supplies five caption segments, each conditioning a block of
$L=21$ latent frames generated in $h=3$-frame chunks. Videos therefore
contain 105 latent frames and 417 decoded frames at $832\times480$
resolution and 16 frames per second, or 26.06 seconds. We use one seed-0
video per case and method. Challenge construction, judge configurations,
and scoring support are detailed in Appendix~\ref{app:online_mbench}.

\paragraph{MBench scores.}
For dimension $a\in\{1,\ldots,6\}$, trigger coverage $\mathrm{C}_a$
measures how often the requested memory challenge occurs. Conditional
reliability $\mathrm{R}_a$ averages valid consistency scores on triggered
cases. We compute
\begin{equation}
    \mathrm{M}_a
    =\frac{2\mathrm{C}_a\mathrm{R}_a}{\mathrm{C}_a+\mathrm{R}_a},
    \qquad
    \overline{\mathrm{M}}=\frac{1}{6}\sum_{a=1}^{6}\mathrm{M}_a.
    \label{eq:exp:mbench-score}
\end{equation}
All scores are on a 0--100 scale. The M-score column in the main table
is $\overline{\mathrm{M}}$, the mean of six dimension-level harmonic
means, not the harmonic mean of average coverage and reliability.

\paragraph{MovieGenBench prompts and generation.}
Following the long-video protocol of Relax Forcing~\citep{relax-forcing},
we use the first 128 MovieGenBench prompts~\citep{moviegen} and five
seeds per prompt, yielding 640 videos per method. Each prompt remains
constant throughout generation. Videos contain 240 latent frames and
957 decoded frames at $832\times480$ and 16 frames per second, or
59.81 seconds. This setting evaluates minute-long generation without
the caption transitions used in MBench.

\paragraph{VBench-Long visual quality.}
We evaluate the six dimensions supported by VBench-Long's custom-input
mode~\citep{vbench,vbench-long}: subject consistency, background
consistency, motion smoothness, aesthetic quality, imaging quality,
and dynamic degree. Subject consistency combines within-clip DINO
features with cross-clip DINOv2 consistency; background consistency
combines within-clip CLIP features with cross-clip DreamSim consistency.
The other dimensions use AMT-S, the LAION aesthetic predictor, MUSIQ,
and RAFT optical flow, respectively.

Each video is divided into 30 fixed two-second clips, with scene-detection
splitting disabled. At 16 frames per second, the final 32-frame clip is
end-aligned and overlaps the preceding clip by three frames. Clip scores
are averaged within each source video and then equally over the 640
videos. Subject and background consistency instead use the evaluator's
fused within-clip and cross-clip source scores. Temporal flickering is
omitted because its static-prompt filter does not apply to this suite.
The column labelled Avg. reports the unweighted mean of the six scores
on the 0--100 scale. It is not the official normalized VBench total and
does not include text alignment.

\subsection{Baseline Policies}
\label{app:baselines}

The main comparisons apply each policy to the same frozen generator
within each backbone. No policy updates the generator weights.
Rolling Sink-style, Deep Forcing-style, and MemRoPE-style adapt published
inference rules rather than use the original methods' complete trained
systems. Active-context budgets differ across policies.

\paragraph{Infinity-RoPE.}
Infinity-RoPE~\citep{yesiltepe_2026_infinityrope} retains one sink frame
and five recent frames under block-relativistic RoPE. The main MBench
and MovieGen evaluations use the released-code default profile without
caption-boundary cache flushing.

\paragraph{Relax Forcing.}
Relax Forcing~\citep{relax-forcing} retains two sink frames, one selected
historical frame, and one recent frame, for $4N$ active past tokens per
layer. The released selector ranks four candidates from the latter half
of the archive using sink similarity minus twice recent-frame similarity.
The first transformer layer selects a shared history index, and each
layer reads its own KV features. The raw-history archive is capped at
120 latent frames.

\paragraph{Rolling Sink-style.}
Our Rolling Sink adaptation~\citep{li_2026_rolling} keeps five sink
blocks and one recent block. With three latent frames per cache block,
this gives 18 cached frames. These cache blocks are native generation
chunks, not PACC's $L$-frame compression blocks.

\paragraph{Deep Forcing-style.}
Our Deep Forcing adaptation~\citep{yi_2025_deep} uses ten sink frames,
four recent frames, and two additional frame-equivalents of participative
compression, for a 16-frame-equivalent budget. Participative compression
retains historical tokens used by recent attention.

\paragraph{MemRoPE-style.}
Our MemRoPE adaptation~\citep{kim_2026_memrope} retains three sink frames,
four local frames, and one long-term and one short-term memory slot formed
by exponential moving averages of past keys. Attention positions are
assigned when the context is read.

\subsection{Detailed Quantitative Results}
\label{app:detailed_results}

We provide the per-dimension results summarized in
Table~\ref{tab:exp_main_comparison}. These are breakdowns of the same
evaluations, not additional experiments. Benchmark definitions and
aggregation procedures are given in Appendix~\ref{app:benchmark-protocols}.

\paragraph{MBench memory dimensions.}
Table~\ref{tab:exp_mbench_main} reports the six individual memory
M-scores. The Human, Object, and Causal columns in the main table average
identity/appearance, geometry/texture, and state progress/progress
correctness, respectively. PACC achieves the highest scores in all six
dimensions on Causal-rCM and five of six on Causal Forcing, where
Infinity-RoPE has the highest texture score.
Judge configurations and scoring-support details are provided in
Appendix~\ref{app:online_mbench}.

\begin{table}[t]
\centering
\caption{\textbf{MBench memory preservation at 26.06 seconds.}
Results use Seed~2.0 triggers and Qwen3-VL-Plus causal-reliability judges,
with one seed-0 video per case and method. Scores are on a 0--100 scale;
M-score is the unweighted mean of the six reported dimensions.
Bold denotes the best value within each backbone.}
\label{tab:exp_mbench_main}
\scriptsize
\setlength{\tabcolsep}{2.8pt}

\resizebox{0.9\linewidth}{!}{%
\begin{tabular}{@{}lccccccc@{}}
\toprule
Method & Identity & Appear. & Geom. & Texture & State & Correct. & M-Score \\
\midrule

\multicolumn{8}{@{}l}{\textit{Causal-rCM c3-3}} \\
Infinity-RoPE~\citep{yesiltepe_2026_infinityrope} & 28.22 & 35.44 & 14.01 & 33.65 & 65.38 & 56.46 & 38.86 \\
Relax Forcing~\citep{relax-forcing} & 25.51 & 28.68 & 24.28 & 48.14 & 64.42 & 67.06 & 43.01 \\
Rolling Sink~\citep{li_2026_rolling} & 13.00 & 18.92 & 3.70 & 5.81 & 45.89 & 38.60 & 20.99 \\
Deep Forcing~\citep{yi_2025_deep} & 27.40 & 35.74 & 21.23 & 47.89 & 52.05 & 46.35 & 38.44 \\
MemRoPE~\citep{kim_2026_memrope} & 28.62 & 35.31 & 11.88 & 50.14 & 64.20 & 56.34 & 41.08 \\
\textbf{PACC (Ours)} & \textbf{36.86} & \textbf{44.34} & \textbf{27.64} & \textbf{54.35} & \textbf{66.58} & \textbf{68.05} & \textbf{49.64} \\
\midrule
\multicolumn{8}{@{}l}{\textit{Causal Forcing}} \\
Infinity-RoPE~\citep{yesiltepe_2026_infinityrope} & 41.85 & 57.98 & 25.85 & \textbf{79.78} & 72.79 & 57.77 & 56.00 \\
Relax Forcing~\citep{relax-forcing} & 43.95 & 59.80 & 23.98 & 77.44 & 74.53 & 71.80 & 58.58 \\
Rolling Sink~\citep{li_2026_rolling} & 45.17 & 55.87 & 17.79 & 72.00 & 68.66 & 55.46 & 52.49 \\
Deep Forcing~\citep{yi_2025_deep} & 36.62 & 54.16 & 24.20 & 79.17 & 70.10 & 54.74 & 53.17 \\
MemRoPE~\citep{kim_2026_memrope} & 34.71 & 54.85 & 25.54 & 78.78 & 74.02 & 60.32 & 54.70 \\
\textbf{PACC (Ours)} & \textbf{49.50} & \textbf{61.90} & \textbf{29.97} & 78.32 & \textbf{76.64} & \textbf{74.27} & \textbf{61.77} \\
\bottomrule
\end{tabular}%
}
\end{table}


\paragraph{VBench-Long quality dimensions.}
Table~\ref{tab:exp_quality} reports the six quality dimensions underlying
the VBench-Long average in the main table. Compared with Relax Forcing,
PACC maintains similar subject and background consistency and motion
smoothness, with higher dynamic degree but lower aesthetic and imaging
scores on both backbones. The aggregate therefore reflects a trade-off
across dimensions rather than uniform quality improvements. The Avg.\
column is the unweighted mean of these six scores, not the official
VBench total.

\begin{table}[t]
\centering
\caption{\textbf{Minute-long video quality on MovieGenBench.}
VBench-Long evaluation on the first 128 prompts with five seeds each
(640 videos per method; 957 frames at 16 fps). Baselines are
inference-policy adaptations on the same frozen backbone. Scores are
on a 0--100 scale; Avg.\ is the unweighted mean of the six dimensions,
not the official VBench total. Bold marks the best value per column
within each backbone; underlining marks the second-best average.}
\label{tab:exp_quality}
\scriptsize
\setlength{\tabcolsep}{2.8pt}

\resizebox{0.9\linewidth}{!}{%
\begin{tabular}{@{}lccccccc@{}}
\toprule
Method & Subject & Backgr. & Motion & Aesthetic & Imaging & Dynamic & Avg. \\
\midrule
\multicolumn{8}{@{}l}{\textit{Causal-rCM c3-3}} \\
Infinity-RoPE~\citep{yesiltepe_2026_infinityrope} & 97.45 & 96.45 & 98.47 & 61.02 & \textbf{70.01} & 57.14 & 80.09 \\
Relax Forcing~\citep{relax-forcing} & 97.13 & 96.21 & 98.35 & 60.84 & 68.83 & 64.64 & \textbf{81.00} \\

Rolling Sink~\citep{li_2026_rolling} & \textbf{98.11} & \textbf{96.95} & \textbf{98.91} & \textbf{61.41} & 69.71 & 39.46 & 77.43 \\
Deep Forcing~\citep{yi_2025_deep} & 97.12 & 96.32 & 98.48 & 59.64 & 67.89 & 60.16 & 79.94 \\
MemRoPE~\citep{kim_2026_memrope} & 97.47 & 96.50 & 98.64 & 60.39 & 69.27 & 55.56 & 79.64 \\
 \textbf{PACC (Ours)}
  & 97.06 & 96.15 & 98.38 & 59.77 & 65.46 & \textbf{66.17} & \underline{80.50} \\
\midrule

\multicolumn{8}{@{}l}{\textit{Causal Forcing}} \\
Infinity-RoPE~\citep{yesiltepe_2026_infinityrope} & 95.55 & 95.25 & 95.89 & 58.70 & 69.65 & 83.23 & 83.04 \\
Relax Forcing~\citep{relax-forcing} & 95.34 & 95.14 & 95.88 & 58.82 & 68.64 & 88.95 & \textbf{83.79} \\

Rolling Sink~\citep{li_2026_rolling} & \textbf{96.33} & \textbf{95.81} & \textbf{97.22} & 58.92 & 69.53 & 75.15 & 82.16 \\
Deep Forcing~\citep{yi_2025_deep} & 94.65 & 94.88 & 95.97 & 55.74 & 66.72 & 90.88 & 83.14 \\
MemRoPE~\citep{kim_2026_memrope} & 95.18 & 95.09 & 96.28 & 57.29 & \textbf{69.79} & 87.69 & 83.55 \\
\textbf{PACC (Ours)}
  & 95.24 & 95.22 & 96.80 & 57.62 & 65.88 & 90.65 & \underline{83.57} \\

\bottomrule
\end{tabular}
}
\end{table}

\subsection{Online MBench Protocol and the Development Subset}
\label{app:online_mbench}

\paragraph{Hosted judges.}
Trigger decisions use Seed~2.0 (\texttt{doubao-seed-2-0-pro-260215});
causal reliability uses Qwen3-VL-Plus
(\texttt{qwen3-vl-plus-2025-12-19}). Both judges receive eight uniformly
sampled decoded frames at temperature zero. The latter replaces the
benchmark's Qwen3-VL-235B evaluator.
Human and object reliability use the released local metrics with corrected
geometry bounds. Generation uses four conditional denoising steps per
$h=3$-frame chunk, seed zero, and one continuous noise stream per case.

\paragraph{Challenge construction.}
MBench builds tests from real long videos drawn from five datasets.
A structured description is divided into five caption segments with
camera-control instructions such as translation, rotation, zoom, or
occlusion. Human and object cases require a subject to appear, leave
view or become occluded, and return. Causal cases require a state change
to progress plausibly even while hidden; 200 base cases yield 391 items
under camera-motion, occlusion, and dimming conditions. Trigger judgments
test whether the challenge occurs, not whether memory is preserved.

\paragraph{Reliability metrics.}
Human identity compares ArcFace face embeddings between the first and
last caption segments; appearance compares DINOv2 features of SAM2 body
masks. For objects, grounding and SAM2 tracking use 32 uniformly sampled
frames. Texture compares tracked DINOv2 features across departure and
return, while geometry measures mask intersection-over-union after
reprojection using Depth Anything~3 depth and camera poses. The causal
judge uses the same eight sampled frames as the trigger judge to assess
state progression and physical plausibility.

\paragraph{Scoring support.}
\label{app:exp_support}
A triggered challenge and a valid reliability measurement are different
quantities: tracking may fail even when the event occurs. Valid track
counts therefore cannot replace the coverage denominator. Reliability
also depends on which cases each policy triggers, so reliability scores
may summarize different event sets. Evaluator failures, including provider
refusals, must be distinguished from negative judgments of generated
content.

\paragraph{Development subset.}
The development manifest contains 137 cases, of which 123 enter the six
reported dimensions: 24 human, 20 object, and 79 causal. This subset is
separate from the full-population main comparison.

\subsection{Native Validation Protocol}
\label{app:exp_training}

\paragraph{Short-rollout validation protocol.}
The native-validation diagnostic uses 12 held-out rollout records and
six prefix lengths per record, with fixed per-case noise streams.
Each checkpoint is evaluated on 72 record--prefix pairs and 1,008
velocity comparisons. We average within each pair and then equally
across pairs, reporting loss on the same $100\times$ velocity-MSE
scale as training. A fixed noise stream does not force different
checkpoints to visit identical generated states.

\subsection{Resources and Evidence Boundaries}
\label{app:exp_cost}

\paragraph{Active context and stored memory.}
The active context in Appendix~\ref{app:pacc-retrieval} remains bounded,
but both memory archives grow with video length. For example, at
$\rho=0.1$, four completed blocks contain $4Q=13{,}104$ memory embeddings. With width
$d=1{,}536$ and BF16 storage, the embeddings occupy approximately
38.4\,MiB. Dense full-width key and value tensors for the same tokens
across 30 layers occupy approximately 2.25\,GiB. These are tensor-size
estimates, not measurements of peak device memory; they omit model
parameters, raw anchors, temporary workspaces, and allocator reserve.
Bounded active context therefore does not by itself imply bounded total
storage or lower end-to-end generation time.

\paragraph{Scope of the comparisons.}
The main results compare complete inference policies on frozen backbones.
Because their context composition and budgets differ, these comparisons
do not isolate the effects of compaction, retrieval, and token count.
Component-level conclusions require the corresponding controlled ablations.

\section{Detailed Ablation Results}
\label{app:ablation_details}

We report the complete subset results underlying
Figure~\ref{fig:ablation_tradeoffs}. The training sweep fixes
$\rho^{-1}=10$ and evaluates 1,000, 2,000, 3,000, 4,000, 5,000,
7,000, and 10,000 updates. The compression sweep fixes 2,000 updates
and evaluates $\rho^{-1}\in\{4,6,8,10\}$. Each VBench-Long setting
contains 64 approximately one-minute MovieGen videos. The shared
tenfold, 2,000-update configuration appears in both sweeps with identical
scores. These subset measurements are distinct from the full main-table
evaluations. We report point estimates without repeated-run or
resampling-based confidence intervals.

\begin{figure}[t]
    \centering
    \includegraphics[width=\linewidth]{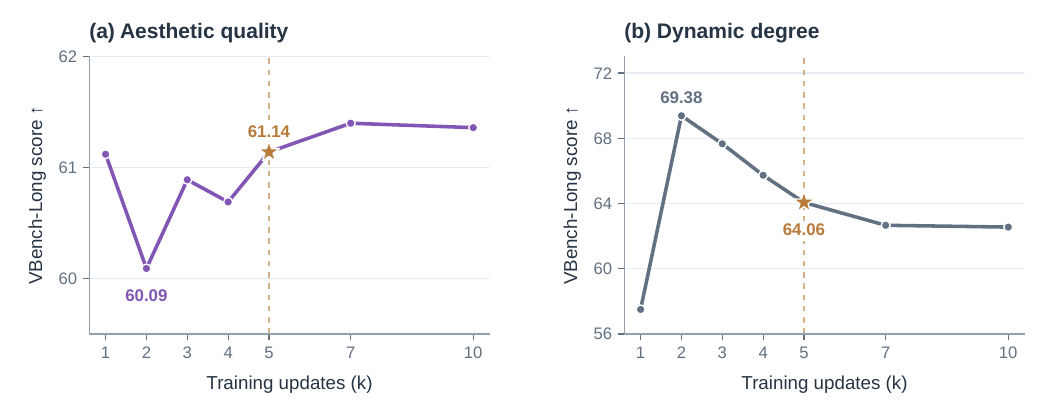}
    \caption{\textbf{Complementary quality dimensions across training.}
    VBench-Long aesthetic quality and dynamic degree at $\rho^{-1}=10$,
    on the same 64-video subset used for imaging quality.
    Stars mark the 5,000-update main checkpoint. Relative to 2,000 updates,
    it has higher aesthetic quality but lower dynamic degree.
    These dimensions are reported separately, not combined into a new score.}
    \label{fig:ablation_quality_details}
\end{figure}

\paragraph{Training duration.}
Tables~\ref{tab:ablation_checkpoints_memory} and
\ref{tab:ablation_checkpoints_quality} report all six memory dimensions
and all six quality dimensions. The 2,000-update checkpoint exceeds
5,000 updates in all six memory dimensions. Conversely, all five quality
dimensions other than dynamic degree have higher point estimates at
5,000 updates. The quality average is slightly lower (80.74 versus
81.01), because the decrease in dynamic degree offsets the increases
elsewhere. Figure~\ref{fig:ablation_quality_details} shows the complementary aesthetic-quality and dynamic-degree results.

The improvement in imaging quality from 2,000 to 5,000 updates does not
establish that 5,000 updates is optimal. For example, the 7,000-update
checkpoint has higher M-score (50.02 versus 47.78) and imaging quality
(68.18 versus 67.21) than the main checkpoint. This illustrates the non-monotonic dependence of memory performance
on training duration.

\begin{table}[t]
\centering
\caption{\textbf{MBench memory in the training-duration sweep at $\rho^{-1}=10$.} Dimension-level M-scores on the MBench evaluation subset. Scores are on a 0--100 scale. M-score is the mean of the six reported memory dimensions. Bold and underlining mark the best and second-best values per column; tied values share their rank. $\dagger$ marks the main checkpoint.}
\label{tab:ablation_checkpoints_memory}
\scriptsize
\setlength{\tabcolsep}{3pt}
\renewcommand{\arraystretch}{1.10}
\resizebox{0.8\linewidth}{!}{%
\begin{tabular}{@{}lrrrrrrr@{}}
\toprule
Updates & Identity & Appear. & Geom. & Texture & State & Correct. & M-score \\
\midrule
1k & 30.55 & 32.99 & 30.24 & \underline{62.06} & 63.27 & 64.46 & 47.26 \\
2k & \textbf{45.12} & \textbf{50.76} & \underline{34.99} & \textbf{66.72} & \underline{69.06} & \textbf{71.31} & \textbf{56.33} \\
3k & 41.05 & 48.44 & 31.86 & 57.26 & \textbf{69.32} & \underline{70.53} & \underline{53.08} \\
4k & \underline{43.83} & \underline{49.75} & 34.44 & 54.38 & 64.79 & 64.95 & 52.02 \\
5k$^{\dagger}$ & 35.24 & 43.30 & 26.01 & 54.13 & 61.09 & 66.92 & 47.78 \\
7k & 40.99 & 38.49 & \textbf{36.26} & 59.03 & 60.54 & 64.79 & 50.02 \\
10k & 34.56 & 34.33 & 34.86 & 54.30 & 61.62 & 58.53 & 46.37 \\
\bottomrule
\end{tabular}%
}
\end{table}

\begin{table}[t]
\centering
\caption{\textbf{VBench-Long quality in the training-duration sweep at $\rho^{-1}=10$.} 64 approximately one-minute MovieGen videos per setting. Scores are on a 0--100 scale. Avg.\ is the unweighted mean of the six quality dimensions, not the official VBench total. Bold and underlining mark the best and second-best values per column; tied values share their rank. $\dagger$ marks the main checkpoint.}
\label{tab:ablation_checkpoints_quality}
\scriptsize
\setlength{\tabcolsep}{3pt}
\renewcommand{\arraystretch}{1.10}
\resizebox{0.8\linewidth}{!}{%
\begin{tabular}{@{}lrrrrrrr@{}}
\toprule
Updates & Subject & Backgr. & Motion & Aesthetic & Imaging & Dynamic & Avg. \\
\midrule
1k & \textbf{97.62} & \textbf{96.40} & \textbf{98.53} & 61.12 & 67.64 & 57.50 & 79.80 \\
2k & 96.99 & 96.04 & 98.34 & 60.09 & 65.21 & \textbf{69.38} & \underline{81.01} \\
3k & 97.32 & 96.20 & 98.25 & 60.89 & 67.18 & \underline{67.66} & \textbf{81.25} \\
4k & 97.28 & 96.20 & 98.36 & 60.69 & 67.12 & 65.73 & 80.89 \\
5k$^{\dagger}$ & 97.41 & 96.28 & 98.35 & 61.14 & 67.21 & 64.06 & 80.74 \\
7k & 97.49 & \underline{96.37} & 98.38 & \textbf{61.40} & \underline{68.18} & 62.66 & 80.74 \\
10k & \underline{97.56} & \underline{96.37} & \underline{98.40} & \underline{61.36} & \textbf{68.38} & 62.55 & 80.77 \\
\bottomrule
\end{tabular}%
}
\end{table}

\paragraph{Compression factor.}
Tables~\ref{tab:ablation_compression_memory} and
\ref{tab:ablation_compression_quality} give the per-dimension results
for the rate sweep in Figure~\ref{fig:ablation_tradeoffs}(c,d).
Tenfold compression has the highest point estimate in all six memory
dimensions, whereas sixfold compression exceeds it on all six quality
dimensions. Thus, the memory advantage does not imply a uniform quality
advantage. Since all rate comparisons are performed at 2,000 updates,
they do not establish a ranking of compression factors at the
5,000-update main checkpoint.

\begin{table}[t]
\centering
\caption{\textbf{MBench memory in the compression-factor sweep at 2,000 updates.} Dimension-level M-scores on the MBench evaluation subset. Scores are on a 0--100 scale. M-score is the mean of the six reported memory dimensions. Bold and underlining mark the best and second-best values per column.}
\label{tab:ablation_compression_memory}
\scriptsize
\setlength{\tabcolsep}{3pt}
\renewcommand{\arraystretch}{1.10}
\resizebox{0.8\linewidth}{!}{%
\begin{tabular}{@{}lrrrrrrr@{}}
\toprule
Factor $\rho^{-1}$ & Identity & Appear. & Geom. & Texture & State & Correct. & M-score \\
\midrule
4 & 32.84 & 36.46 & \underline{32.21} & 61.99 & \underline{64.38} & \underline{69.77} & \underline{49.61} \\
6 & 29.23 & 31.60 & 29.28 & \underline{62.52} & 62.38 & 63.39 & 46.40 \\
8 & \underline{38.57} & \underline{39.63} & 27.32 & 55.18 & 61.59 & 62.74 & 47.51 \\
10 & \textbf{45.12} & \textbf{50.76} & \textbf{34.99} & \textbf{66.72} & \textbf{69.06} & \textbf{71.31} & \textbf{56.33} \\
\bottomrule
\end{tabular}%
}
\end{table}

\begin{table}[t]
\centering
\caption{\textbf{VBench-Long quality in the compression-factor sweep at 2,000 updates.} 64 approximately one-minute MovieGen videos per setting. Scores are on a 0--100 scale. Avg.\ is the unweighted mean of the six quality dimensions, not the official VBench total. Bold and underlining mark the best and second-best values per column.}
\label{tab:ablation_compression_quality}
\scriptsize
\setlength{\tabcolsep}{3pt}
\renewcommand{\arraystretch}{1.10}
\resizebox{0.8\linewidth}{!}{%
\begin{tabular}{@{}lrrrrrrr@{}}
\toprule
Factor $\rho^{-1}$ & Subject & Backgr. & Motion & Aesthetic & Imaging & Dynamic & Avg. \\
\midrule
4 & \textbf{97.51} & \textbf{96.46} & \textbf{98.43} & \textbf{61.64} & \underline{68.29} & 62.40 & 80.79 \\
6 & \underline{97.41} & \underline{96.28} & 98.35 & \underline{61.34} & \textbf{68.43} & \textbf{70.52} & \textbf{82.06} \\
8 & 97.29 & 96.23 & \underline{98.42} & 60.63 & 65.64 & 66.61 & 80.80 \\
10 & 96.99 & 96.04 & 98.34 & 60.09 & 65.21 & \underline{69.38} & \underline{81.01} \\
\bottomrule
\end{tabular}%
}
\end{table}

\section{Additional Qualitative Results}
\label{app:qualitative_results}

\begin{figure}[t]
    \centering
    \includegraphics[width=\linewidth]{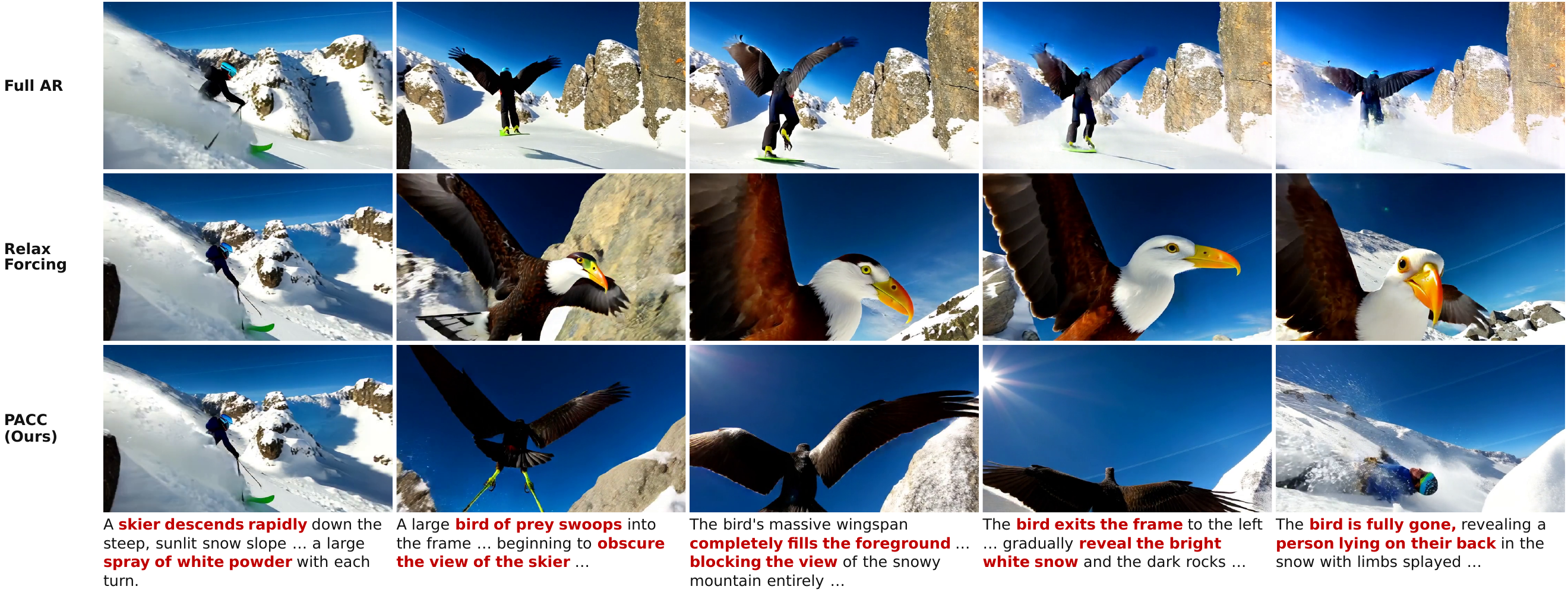}
    \caption{\textbf{Qualitative memory comparison on Causal-rCM c3-3.}
    Columns follow five consecutive prompt segments, with excerpts shown
    below. A bird passes in front of a skier, obscuring the scene before
    leaving to reveal the skier on the snow. PACC depicts this
    occlusion-and-reveal sequence, whereas Full AR and Relax Forcing
    do not show the requested final reveal in the displayed frames.}
    \label{fig:qualitative_memory_rcm}
\end{figure}

Figure~\ref{fig:qualitative_memory_rcm} complements the Causal Forcing
example in Figure~\ref{fig:qualitative_memory} with a comparison on
Causal-rCM under foreground occlusion.

\end{document}